# REaL: Real-time Face Detection and Recognition Using Euclidean Space and Likelihood Estimation

Sandesh Ramesh, Manoj Kumar M V, K Aditya Shastry
Department of Information Science and Engineering
Nitte Meenakshi Institute of Technology, Bangalore
Sandeshr1515@gmail.com, manoj.kumar@nmit.ac.im,
adityashastry.k@nmit.ac.in

**Abstract.** Detecting and recognizing faces accurately has always been a challenge. Differentiating facial features, training images, and producing quick results require a lot of computation. The REaL system we have proposed in this paper discusses its functioning and ways in which computations can be carried out in a short period. REaL experiments are carried out on live images and the recognition rates are promising. The system is also successful in removing non-human objects from its calculations. The system uses a local database to capture images and feeds the neural network frequently. The captured images are cropped automatically to remove unwanted noise. The system calculates the Euler angles and also the probability of whether the face is smiling, has its left eye, and right eyes open or not.

**Keywords –** Face detection, recognition, security, computer vision, neural network, images

## I. INTRODUCTION

It is said that for every person, there are seven look-alikes all around the world. But even these look-alikes aren't completely alike. While the fine details are oblivious to the naked eye from afar, upon close inspection, the differences are many and diverse. Let us also consider the case of monozygotic twins. These twins have the same genetic information. They even look the same. For a stranger, it wouldn't require much effort to go wrong. But these twins, too, have differentiating features, and these features are visible right on their faces. Earlier in the days, twins were considered a rarity. But in modern times, having twins is not surprising, given our way of life. Some of the factors that influence mothers to give birth to twins are as follows –

1. Age of mother - Generally, women in their 40s tend to produce higher levels of oestrogen than their younger counterparts. This means that their ovaries produce more than one egg at a given time [1].
2. Pregnancy count – Greater the number of times a woman becomes pregnant, the greater the chances of having twins.
3. Genetics – If the mother is a twin herself, the chances she's going to

conceive twins are high.

Under any of these circumstances, each baby is born unique in its own way. Each one of us is unique. We wake up every day to look at ourselves in the mirror and do everything we can to look good. We value our faces a lot. But with the evolution of technology, no one ever really thought that their face was going to be their password to unlock their phones or bypass security checks at the airport. With rapid advancements in technology, let alone individuals, twins can be differentiated in seconds. So how does this work? Various algorithms have been developed that can form patterns on our face, capture our retinal design or mimic our entire face in 3D. The system we have developed focuses on capturing the facial data through a mobile application, which in turn can be used for surveillance without much effort. The system is also designed to name the person in the image to assure the user of its accuracy. If the system is unable to recognize the person, the system automatically begins to capture images and train them using neural networks. The upcoming sections are ordered in the following pattern – Section II focuses on previous research work conducted by those researching on facial recognition technology. Section III illustrates an overview of our recognition system REaL. Section IV illustrates the methodology of REal approaches. Section V tabulates the results and readings obtained by the system. Section VI examines the future open research opportunities in facial recognition. Section VII concludes with a concise summary of the entire paper.

## II. LITERATURE SURVEY

In the paper titled "Robust Real-Time Face Detection", the authors have proposed a framework for face detection that is capable of processing images at a very fast pace. The entire functioning of the framework is based on three contributions – the first is the introduction of "Integral Image". With this, the detectors feature can perform faster computations. The second contribution involves an efficient classifier which is designed to capture only certain important facial features from a bag of many features. The third contribution involves the removal of background images to focus more on computing important facial features [2]. Oliver Jesorsky et. al., in their paper titled "Robust Face Detection Using the Hausdorff Distance", have proposed an approach of shape comparison to obtaining quick and accurate detection of faces in still greyscale images only. The Hausdorff Distance acts as a similarity model between faces in an image and the objects present in it. The approach accommodates a dual step process to remove the coarse and detect the exact location of a face in an image [3]. Research paper titled "Real-time Face Detection on a Configurable Hardware System", discusses a real-time face detection method and its implementation on configurable hardware. The system functions at 30 frames per second and uses a neural network to filter the input images [4]. According to the author, only a single operation is sufficient to accurately detect faces. G. Krishna et. al., in their paper "Face Detection System on AdaBoost Algorithm Using Haar Classifiers", have utilized AdaBoost algorithm using Haar features to detect faces. Techniques such as image scaling, parallel processing, and integral image generation have been used to speed up the process of face detection [5]. In the paper titled "Simple and Fast Face Detection System Based on Edges", the authors propose a three-step approach to detect faces easily and quickly. The first step involves the removal of noise from the images by using filters. The second step, using the Sobel operator, edge images are constructed and enhanced. The third step involves the application of an edge tracking algorithm to enhance the images based on edges [6]. In their paper "Face Detection System Based on Feature-Based Chrominance Color Information", the authors have proposed a system wherein chrominance colour information is extracted from an image containing a face and its background. The method primarily focuses on detecting the eyes, brows, nose, and mouth. Their experimental results produced a high accuracy for detecting frontal part of the face in very less time [7]. Research paper titled "Face Detection Using Representative Learning", discusses ways to detect faces fast, using convolutional neural network (CNN) by explicitly capturing facial features. They propose that their model's CNN can train images extensively using Adaboost algorithms by extracting facial features [8].

## III. FRAMEWORK

The main purpose of the REaL system is to detect and recognize faces in a short period. The system works well on a smartphone that has a camera of moderate to high quality. If the environment is dimly lit, the phone's flashlight can be used to assist the system. We have utilized Google's Firebase Face Detection API into our system. When the system begins to capture faces, the system concurrently runs through its database to find a match. If a match has been identified, the same is outputted side-by-side the captured image along with the ID assigned to it. If the system doesn't recognize a face, it prompts the user whether to capture and store the images or not. If yes, the system begins to capture images at 60 frames per second and stores them in the database (in the case of our research, our phone was used as the local database). Once the images have been captured, the system automatically allocates a unique ID. The system also prompts the user if they'd like to name the face. If yes, a name can to type in. If not, the system displays the ID of the face the next time it detects it. It is important to surround the face with a bounding box. This helps in cropping the image to the maximum extent and thereby reducing storage space while using these cropped images to run neural network models for training.

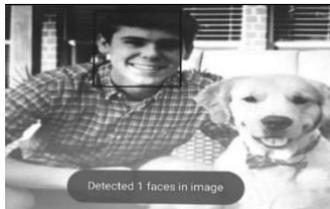

*Fig.1. REaL recognises only human faces*

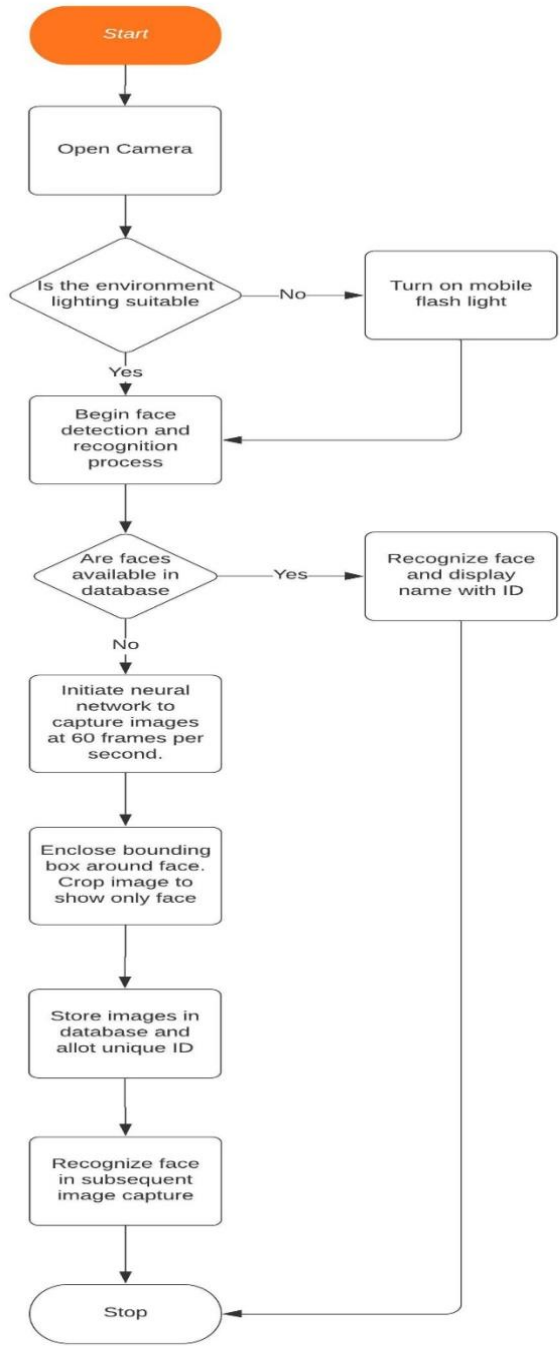

*Fig. 2. Course taken by REal System*

# IV. METHODOLOGY

The table illustrated in the subsequent section contains the following abbreviations:
1. Zzwy – getBoundingBox() – Returns the axis-aligned bounding rectangule of the detected face.
2. Zzxq – getTrackingId() – Returns the tracking ID if tracking is enabled.
3. Zzxu – getHeadEulerAngleY() – Returns the rotation of the face about the vertical axis of the image.
4. Zzxv – getHeadEulerAngleZ() – Returns the rotation of the face about the vertical axis of the image.
5. Zzxt – getIsSmilingProbability() – Returns a value between 0.0 and 1.0 giving a probability that the face is smiling.
6. Zzxs – getIsLeftEyeOpenProbability() – Returns a value between 0.0 and 1.0 giving a probability that the left eye is open.
7. Zzxr – getIsRightEyeOpenProbability() – Returns a value between 0.0 and 1.0 giving a probability that the right eye is open.

# V. RESULTS

For experimentation, tracking ID Zzxq is taken as 0. The system successfully recognises the face and creates a bounding box (as represented by Zzwy). The system also detects whether the face is smiling, has its left eye open, and right eye open or not. Imaginary axis created detects the angle at which the face is placed. This is represented by Zzxv and Zzxu for Z and Y-axis respectively. The trial is conducted for an unidentified face.

TABLE 1. QUANTITATIVE ATTRIBUTES OF REaL FACE RECOGNITION SYSTEM

| Observation | Image | Zzwy | Zzxq | Zzxu | Zzxv | Zzxt | Zzxs | ZZxr |
|---|---|---|---|---|---|---|---|---|
| 1. | 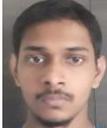 | Rect(290,467-389,718) | 0 | -4.8975 | -3.3457 | 0.0 | 1.0 | 1.0 |
| 2. | 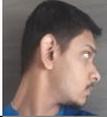 | Rect(127,436-225,667) | 0 | -2.6786 | -0.7822 | 0.0 | 0.0 | 1.0 |
| 3. | 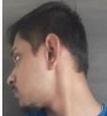 | Rect(321,478-311,553) | 0 | -3.6663 | -2.7923 | 0.0 | 1.0 | 0.0 |

| 4. | 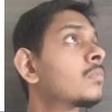 | Rect(288,499-411,589) | 0 | -0.7819 | -2.8871 | 0.0 | 0.0 | 1.0 |
|---|---|---|---|---|---|---|---|---|
| 5. | 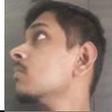 | Rect(324,467-231,784) | 0 | -2.1657 | -1.9878 | 0.0 | 1.0 | 0.0 |
| 6. | 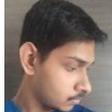 | Rect(123,345-487,811) | 0 | -4.9288 | -2.1289 | 0.0 | 0.0 | 1.0 |
| 7. | 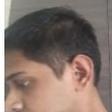 | Rect(215,488-376,899) | 0 | -5.2213 | -2.6616 | 0.0 | 1.0 | 0.0 |
| 8. | 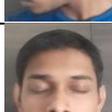 | Rect(118,443-486,677) | 0 | -3.8987 | -2.7618 | 0.0 | 0.0 | 0.0 |
| 9. | 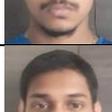 | Rect(288,314-420,599) | 0 | -4.1278 | -4.8762 | 1.0 | 1.0 | 1.0 |
| 10. | 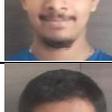 | Rect(301,389-473,784) | 0 | -3.8928 | -4.9921 | 1.0 | 1.0 | 1.0 |

## VI. FUTURE SCOPE

The system we have designed can identify and recognise only a limited number of faces at a time. The system also takes time to compute readings, store images and attributes, and display the output. With the introduction of a more powerful neural network capable of computing hundreds of images at a time, the entire process of surveillance becomes easy and quick. The real challenge is to identify faces in a crowd that is very large and compact [9]. If the system identifies an unrecognized face it must capture enough images in a fraction of a second and train them immediately. Should a solution be found for these problems, the process of recognition could be very successful and reliable.

## VII. CONCLUSION

Through this paper, we have successfully demonstrated the use of our system REaL in capturing live images and recognising them quickly. Security is at an all-time high is a lot of places. With the use of technology, and with the advancements in computer vision and computer security, products and applications such as REaL can be built to protect and monitor people. While this system is not just limited to surveillance, it can be used as a passcode for carrying out transactions without the use of PINs or passcodes. With rapid advancements in CV, future transactions could run through using only our faces.